\documentclass[sigconf]{acmart}
\usepackage[table]{xcolor}
\usepackage{graphicx}
\usepackage{balance}
\AtBeginDocument{%
  }

\setcopyright{acmlicensed}
\copyrightyear{2018}
\acmYear{2018}
\acmDOI{XXXXXXX.XXXXXXX}
\acmConference[Conference acronym 'XX]{Make sure to enter the correct
  conference title from your rights confirmation email}{June 03--05,
  2018}{Woodstock, NY}
\acmISBN{978-1-4503-XXXX-X/2018/06}

\begin{document}
\title{Traits Run Deeper: Trait-Specific Asymmetric Fusion for Multimodal Personality Assessment}

\author{Jia Li}
\authornote{Corresponding author. }
\affiliation{%
  \institution{Hefei University of Technology}
  \city{Hefei}
  \country{China}
}
\email{jiali@hfut.edu.cn}

\author{Qian Chen}
\affiliation{%
  \institution{Hefei University of Technology}
  \city{Hefei}
  \country{China}
}
\email{cqian@mail.hfut.edu.cn}

\author{Wei Wang}
\affiliation{%
  \institution{Hefei University of Technology}
  \city{Hefei}
  \country{China}
}
\email{2024218391@mail.hfut.edu.cn}

\author{Xinyu Li}
\affiliation{%
  \institution{Hefei University of Technology}
  \city{Hefei}
  \country{China}
}
\email{lixinyu@mail.hfut.edu.cn}

\author{Zhenzhen Hu}
\affiliation{%
    \institution{Intelligent Interconnected Systems Laboratory of Anhui Province}
    \institution{Hefei University of Technology}
    \city{Hefei}
    \country{China}
}
\email{zzhu@hfut.edu.cn}

\author{Dongsheng Shao}
\affiliation{%
  \institution{Jianghuai Advanced Technology Center (Jianghuai Lab)}
  \institution{Anhui Provincial Industry Innovation Center of Humanoid Robots}
  \institution{Anhui Provincial Key Laboratory of Humanoid Robots}
  \city{Hefei}
  \country{China}
}
\email{sds20060021@126.com}

\author{Richang Hong}
\affiliation{%
  \institution{Hefei University of Technology}
  \city{Hefei}
  \country{China}
}
\email{hongrc.hfut@gmail.com}

\author{Meng Wang}
\affiliation{%
  \institution{Hefei University of Technology}
  \city{Hefei}
  \country{China}
}
\email{eric.mengwang@gmail.com}

\renewcommand{\shortauthors}{Jia Li et al.}

\begin{abstract}

Personality assessment aims to infer stable traits from dynamic behaviors across modalities like language, voice, and facial expressions. Existing approaches often adopt a uniform multimodal fusion strategy for all personality dimensions, overlooking trait-specific modality preferences and causing cross-modal interference. To address this, we propose \textbf{Traits Run Deeper}, a novel personality assessment framework consisting of three components. First, the \textbf{Multimodal Foundation Representation (MFR)} module constructs personality-oriented inputs and incorporates psychology-informed semantic templates as anchors, enabling foundation models to capture trait-relevant behaviors. Second, the \textbf{Trait-Specific Modality Fusion (TSMF)} module employs an asymmetric fusion mechanism, allowing each dimension to selectively exploit different modality pathways to capture heterogeneous preferences while reducing cross-modal contamination. Third, the \textbf{Distribution-Calibrated Personality Regression (DCPR)} module mitigates label imbalance and central tendency bias through target distribution calibration, improving robustness and stability. Experimental results on the AVI Challenge 2026 validation set show that our framework reduces mean squared error (MSE) by approximately 25\% compared with the baseline. Consistent improvements on the official test set demonstrate that our method achieves the best performance and ranks first in the AVI Challenge 2026 Personality Assessment Track. The source code will be made available at \url{https://github.com/MSA-LMC/TraitsRunDeeper}.

\end{abstract}



\maketitle

\section{Introduction}
Personality assessment is crucial for understanding individual behavioral styles and psychological tendencies in human-centered applications, such as teamwork and interview analysis~\cite{Barrick1991THEBF, Paunonen2001BigFF, 10.1109/TAFFC.2015.2513401}. Driven by deep learning and multimodal perception, personality recognition has shifted from questionnaire evaluations to automatic analysis from natural behavioral responses. While early benchmarks focused on apparent Big-Five prediction from short videos~\cite{PonceLpez2016ChaLearnL2, Zhang2016DeepBR}, interview-oriented tracks like the AVI Personality Assessment Challenge require participants to infer personality scores from trait-related responses~\cite{Escalante2017DesignOA, 10.1145/3746027.3762016}. This setting demands capturing deeper trait-related cues from limited multimodal evidence.

\begin{figure}[!t]
  \centering
  \includegraphics[width=\linewidth]{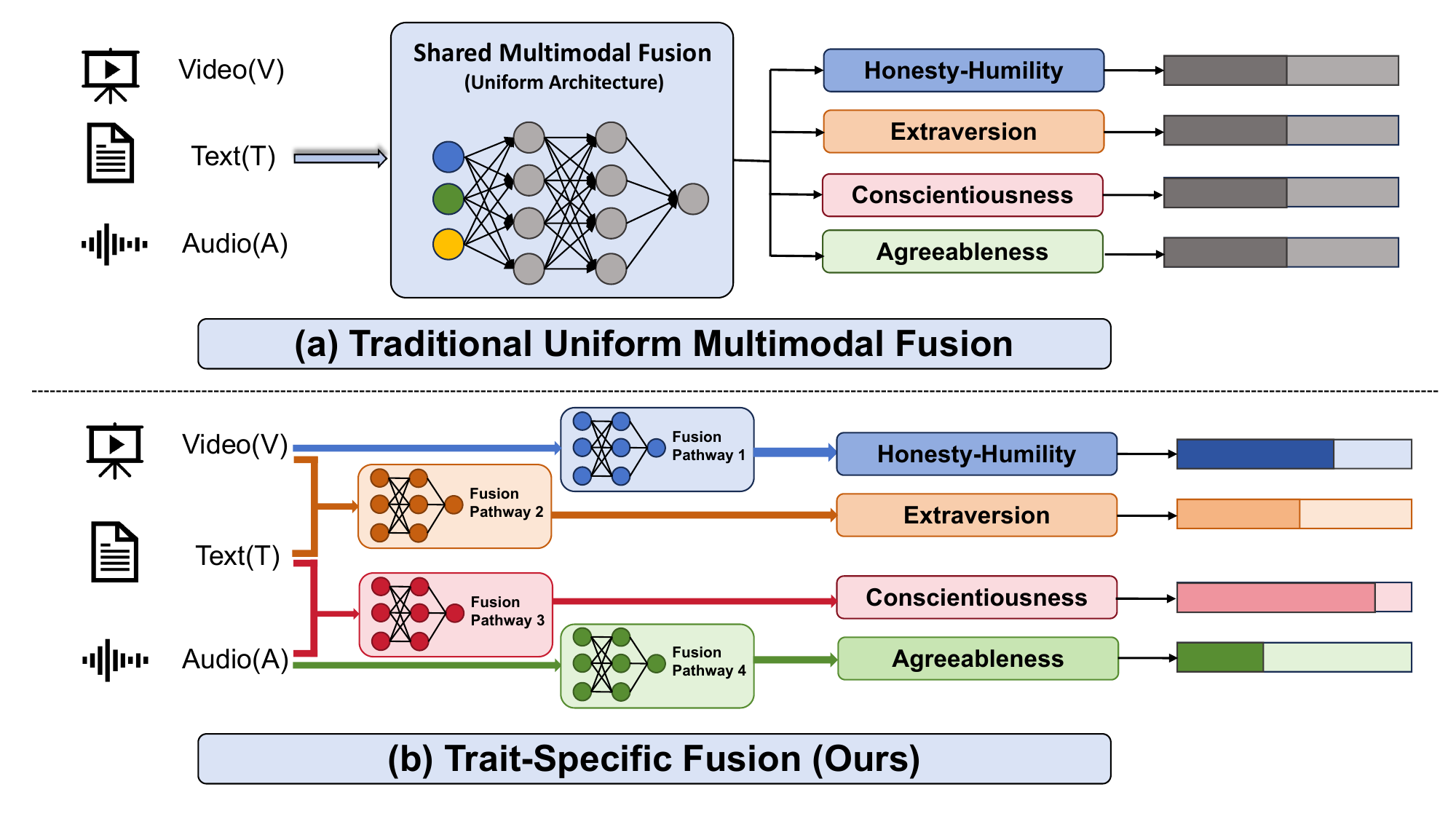}
  \caption{\textbf{Comparison of multimodal fusion paradigms for personality assessment.}}
  \label{fig:motivation}
\end{figure}

As illustrated in Figure~\ref{fig:motivation}(a), existing methods typically extract audio, video and text representations using pretrained models and adopt a uniform prediction structure for all traits~\cite{Aslan2021MultimodalAO, 10.1109/TAFFC.2024.3363710}. However, psychological research indicates that different personality dimensions are revealed through distinct behavioral channels rather than being uniformly distributed. According to Brunswik's lens model~\cite{Vinciarelli2014ASO}, socially-oriented and emotionally expressive traits (e.g., Extraversion or Agreeableness) leak heavily through nonverbal cues like facial expressions, body gestures, and vocal variations~\cite{10.5555/1622637.1622649}. Conversely, traits reflecting cognitive processing or self-organization (e.g., Conscientiousness or Neuroticism) are more robustly manifested in linguistic structures and semantic consistency~\cite{Pennebaker1999LinguisticSL}. Overlooking these trait-specific modality preferences can introduce irrelevant cross-modal noise and cause severe cross-modal contamination under small-sample conditions.

To address these limitations, we propose a Trait-Specific Asymmetric Fusion framework. First, Multimodal Foundation Representation (MFR) module extracts features across modalities using foundation models~\cite{Lee2024GeckoVT}, incorporating psychology-informed semantic templates as anchors to capture trait-relevant information. As conceptualized in Figure~\ref{fig:motivation}(b), TSMF formulates trait-specific modality utilization as a validation-guided model selection process, allowing each personality dimension to select the modality combination and fusion pathway that best matches its behavioral characteristics.~\cite{tan-bansal-2019-lxmert}. Finally, Distribution-Calibrated Personality Regression (DCPR) module applies Yeo-Johnson transformation~\cite{10.1093/biomet/87.4.954} and Gaussian smoothing to calibrate target imbalances. A five-fold regression ensemble is further adopted to enhance stability.

Our framework was evaluated on the AVI Challenge 2026 dataset. Five-fold cross-validation yields an average MSE of 0.2512 across the four personality dimensions. On the official test set, our method obtains an MSE of 0.27767.

\section{Related Works}

\subsection{Automatic Personality Assessment}
Traditional personality assessment relies on questionnaires like the Five-Factor and HEXACO models~\cite{john1999big,ashton2009hexaco}, which are theoretically grounded but suffer from self-report bias, low efficiency, and limited scalability. Consequently, automatic personality recognition has emerged to study observable behavioral signals~\cite{Vinciarelli2014ASO,Zhao2022DeepPT}. While early benchmarks used short videos (e.g., ChaLearn First Impressions) or dyadic scenarios (e.g., UDIVA)~\cite{ponce2016chalearn,Peters2023LargeLM}, recent studies incorporate deep learning, multimodal representations, LLM embeddings, and psychology-guided or LLM-enhanced fusion~\cite{li2025traits,cui2025less,yang2025enhancing,Peters2023LargeLM}. The AVI Challenge 2026 further emphasizes robust assessment from structured interviews~\cite{zhangavi}. However, existing methods mostly focus on overall accuracy and largely overlook the heterogeneous modality preferences of different personality traits.

\subsection{Multimodal Representation and Fusion}
Multimodal personality assessment leverages visual, acoustic, and textual modalities. To capture complementary cues, studies increasingly employ foundation models and deep representation learning to extract high-level features~\cite{zhao2023integrating,ryumina2024gated,cheng2025vaemo}, where LLM embeddings and multimodal foundation representations have shown strong effectiveness~\cite{cui2025less,li2025traits,yang2025enhancing}. Existing exploration spans various paradigms, including attention-based interaction~\cite{electronics14142837}, temporal facial modeling~\cite{Wang2024ASM,11207542}, multi-scale or segment fusion~\cite{Ning2026ADM, 10.1145/3746027.3762064}, alongside early, late, gated, and hierarchical fusion architectures~\cite{ryumina2024gated,masumura2025multimodal}. Although affective computing research suggests that different targets benefit from label-aware fusion and unique modality cues~\cite{Pham2018FoundIT,Zhang2022TailorVM, LI2025113260, qi2026affectspeech}, most personality assessment systems still adopt a shared fusion structure across all dimensions, which introduces irrelevant cross-modal noise and weakens trait-specific behavioral evidence.

\subsection{Distribution-Aware Regression and Label Calibration}
Beyond fusion, the distribution of supervision signals critically impacts regression. Personality and emotional annotations are often imbalanced, noisy, and concentrated around central values~\cite{8913501, Park2020KEmoConAM}, hindering model optimization. To alleviate this, recent works explore label distribution learning, target smoothing, and imbalance-aware regression~\cite{Ren2022BalancedMF,Zhu2024IRDAID,Puetz2026DeconstructingDI}. Statistically, target transformations like Yeo-Johnson are widely adopted to reduce skewness and stabilize variance~\cite{10.1093/biomet/87.4.954}. However, such distribution calibration strategies remain underexplored in multimodal personality assessment, motivating the integration of distribution-aware target calibration in our framework.

\section{Method}
Our research focuses on predicting continuous scores for four dimensions of the HEXACO personality model, including Honesty-Humility, Extraversion, Agreeableness, and Conscientiousness. To better exploit the relationship between personality traits and multimodal behavioral evidence, we propose the \textbf{Traits Run Deeper} framework. The framework consists of Multimodal Foundation Representation (MFR), Trait-Specific Modality Fusion (TSMF), and Distribution-Calibrated Personality Regression (DCPR). In the textual stream, psychology-informed semantic anchors are incorporated before embedding extraction to emphasize personality-relevant cues and improve textual representation quality.

\begin{figure*}[t]
    \centering
    \includegraphics[width=\textwidth]{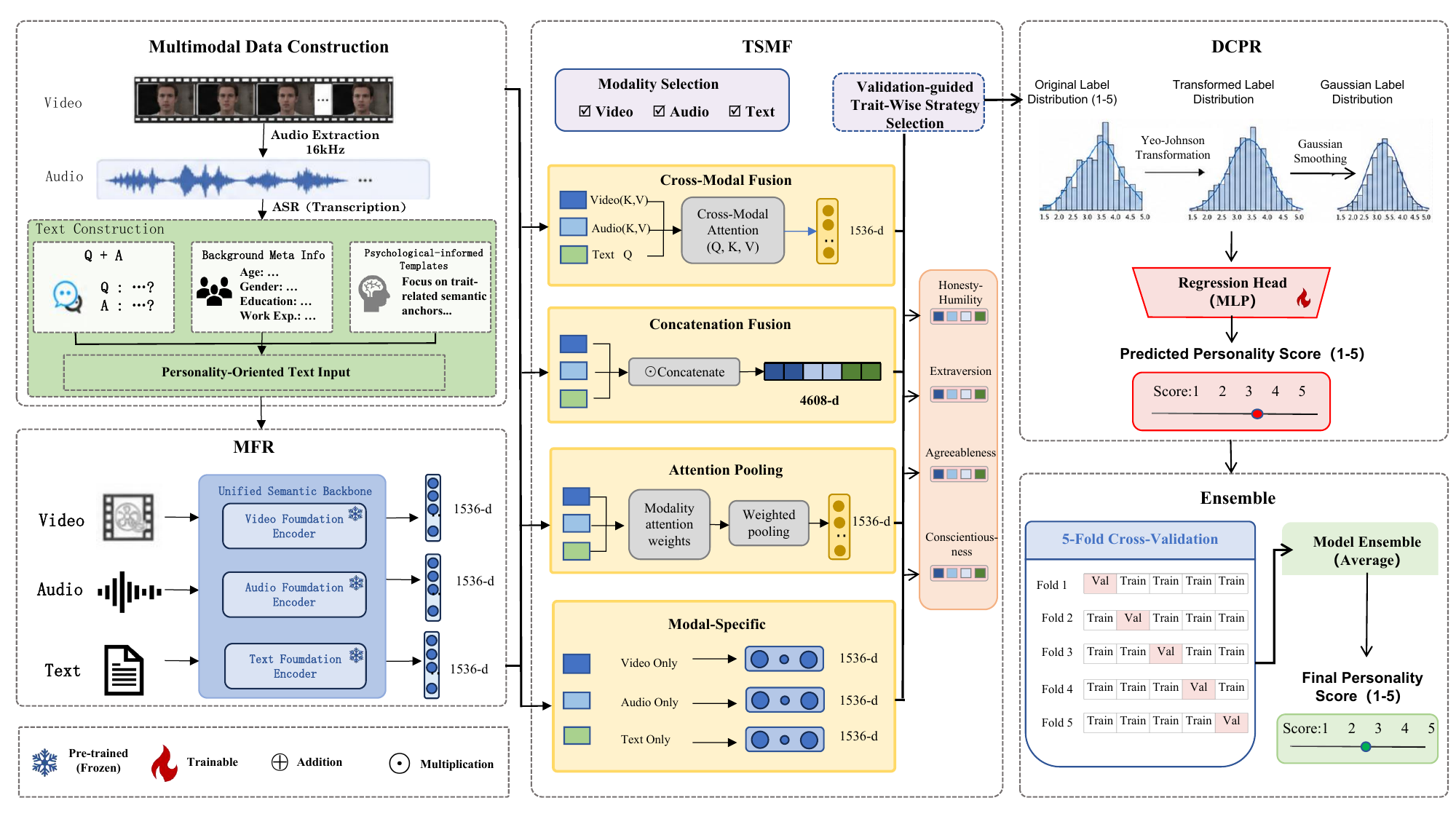}
    \caption{Overview of the proposed Traits Run Deeper framework. The framework consists of three stages: Multimodal Foundation Representation (MFR), Trait-Specific Modality Fusion (TSMF), and Distribution-Calibrated Personality Regression (DCPR). Different personality traits are allowed to select distinct modality combinations and fusion strategies according to their modality preferences.}
    \label{fig:architect}
\end{figure*}

\subsection{Multimodal Data Construction and Foundation Representation}

The AVI Challenge provides raw interview videos as input. We therefore first construct three modalities, including video, audio, and text, for personality assessment. For video, original recordings are directly input without face-centric preprocessing to preserve complete visual context and capture broader behavioral cues. For audio, speech tracks are extracted via FFmpeg \footnote{\url{https://ffmpeg.org/}} and resampled to 16\,kHz.

To construct the text modality, the extracted speech is transcribed using the Whisper-small~\cite{pmlr-v202-radford23a} model. Since raw transcripts may not explicitly emphasize personality-related semantics, we further construct a personality-oriented textual description by integrating three components: (1) question-answer content, (2) demographic metadata provided by the challenge organizers, and (3) psychology-informed semantic templates derived from personality theory. Rather than for instruction following or explicit personality reasoning, these templates serve as semantic anchors introducing trait concepts. Specifically, each text input is prefixed with a prompt formulation: \textit{``Analyze the transcript by focusing on [Trait Anchors].''} For instance, the Extraversion template inserts concepts like \textit{``social engagement and communication style,''} while the Conscientiousness template introduces \textit{``organization, responsibility, and planning.''} In practice, we design multiple template variants with different levels of psychological specificity and select the final template set according to validation performance.

After multimodal data construction, foundation representation models are used to encode video, audio, and text independently. We investigate several candidate representation models, including conventional modality-specific models and recent multimodal embedding models, with detailed comparisons presented in Section~\ref{sec:experiments}. Gemini Embedding 2~\cite{shanbhogue2026geminiembedding2native} is employed as the default backbone and generates a 1536-dimensional representation for each modality. As illustrated in Figure~\ref{fig:architect}, the resulting multimodal representations are subsequently used for trait-specific personality modeling.

\subsection{Trait-Specific Modality Fusion}

After obtaining video, audio, and text representations, we construct a trait-specific fusion space for personality prediction. The key idea is to formulate trait-specific modality utilization as a validation-guided model selection problem rather than enforcing a shared multimodal architecture. Guided by the psychological hypothesis that different personality traits rely on different behavioral channels, each trait independently selects the modality combination and fusion strategy with the best validation performance. The final modality-fusion configuration for each trait is determined through validation-based model selection, while the candidate design is shared across traits.

Denote the extracted video, audio, and text representations as $\mathbf{z}_v$, $\mathbf{z}_a$, and $\mathbf{z}_t$, respectively. For the $k$-th personality trait, a candidate modality set is represented as
\begin{equation}
\mathcal{S}_k \subseteq \{v,a,t\}.
\end{equation}
Given $\mathcal{S}_k$, a fusion function $g_k(\cdot)$ is applied to produce the trait representation:
\begin{equation}
\mathbf{u}_k = g_k(\{\mathbf{z}_m \mid m \in \mathcal{S}_k\}),
\end{equation}
where $\mathbf{u}_k$ is used for predicting the $k$-th personality score. In this formulation, different traits can be associated with different modality evidence and different fusion functions.

For single-modality configurations, no cross-modal fusion is required. The selected representation is directly used as the trait feature:
\begin{equation}
\mathbf{u}_k = \mathbf{z}_{m_k}, \quad m_k \in \{v,a,t\}.
\end{equation}
This pathway is included because some personality traits may be better captured by one dominant modality than by a forced multimodal combination.

For multimodal configurations, we first consider concatenation fusion. Given a selected modality set $\mathcal{S}_k$, the corresponding representations are concatenated and projected into the regression space:
\begin{equation}
\mathbf{u}_k =
\phi_{\mathrm{cat}}
\left(
[\mathbf{z}_m]_{m \in \mathcal{S}_k}
\right),
\end{equation}
where $\phi_{\mathrm{cat}}(\cdot)$ denotes a learnable projection function. This pathway preserves all selected modality features and serves as a simple but strong multimodal baseline.

We further introduce attention pooling to learn the relative importance of selected modalities. For each modality $m \in \mathcal{S}_k$, its attention weight is computed as
\begin{equation}
\alpha_m =
\frac{
\exp(\mathbf{w}^{\top}\tanh(W\mathbf{z}_m))
}{
\sum_{j \in \mathcal{S}_k}
\exp(\mathbf{w}^{\top}\tanh(W\mathbf{z}_j))
},
\end{equation}
and the fused representation is obtained by
\begin{equation}
\mathbf{u}_k =
\sum_{m \in \mathcal{S}_k}
\alpha_m \mathbf{z}_m .
\end{equation}
Compared with direct concatenation, attention pooling allows the model to assign different weights to modalities within the selected combination.

For traits whose selected modality set contains text and at least one non-text modality, we also consider text-centered cross-modal attention. This design uses the text representation as the query, while the non-text representations provide keys and values. Specifically,
\begin{equation}
\mathbf{Q} = W_q \mathbf{z}_t,
\quad
\mathbf{K} =
W_k [\mathbf{z}_m]_{m \in \mathcal{S}_k \setminus \{t\}},
\quad
\mathbf{V} =
W_v [\mathbf{z}_m]_{m \in \mathcal{S}_k \setminus \{t\}} .
\end{equation}
The cross-modal context is computed as
\begin{equation}
\mathbf{c}_k =
\mathrm{softmax}
\left(
\frac{\mathbf{Q}\mathbf{K}^{\top}}{\sqrt{d}}
\right)
\mathbf{V},
\end{equation}
and the final representation is obtained by combining the textual feature with the attended non-text context:
\begin{equation}
\mathbf{u}_k =
\phi_{\mathrm{cm}}
([\mathbf{z}_t;\mathbf{c}_k]).
\end{equation}
This one-way attention structure avoids unnecessary bidirectional interaction and uses text as the semantic anchor for aligning audio or video evidence.

Finally, each personality dimension is predicted by an independent regression branch:
\begin{equation}
\hat{y}_k = r_k(\mathbf{u}_k),
\end{equation}
where $r_k(\cdot)$ denotes the regression head for the $k$-th trait. The modality combination and fusion strategy are selected separately for each trait according to validation performance, and the resulting configuration is analyzed in the experimental section.

\subsection{Distribution-Calibrated Personality Regression}

\begin{figure}[t]
\centering
\includegraphics[width=\linewidth]{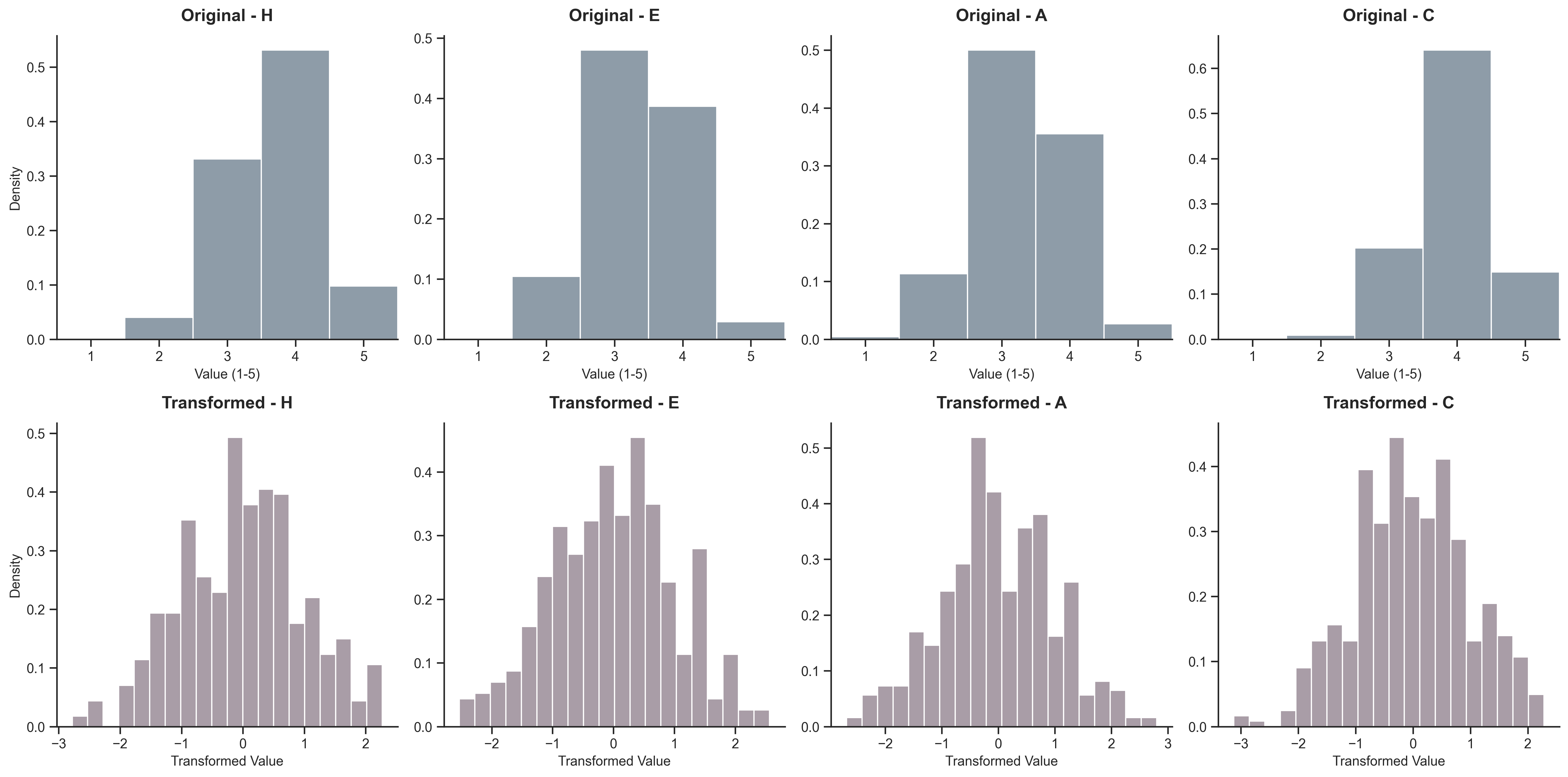}
\caption{
Distribution of personality scores before and after the calibration.
}
\label{fig:label_dist}
\end{figure}

Personality annotations in the AVI Challenge are obtained through self-assessment questionnaires and exhibit noticeable distribution imbalance and central tendency bias. As shown in Figure~\ref{fig:label_dist}, most samples are concentrated around moderate score ranges, while extreme personality scores occur much less frequently. Such characteristics may bias model optimization and hinder the learning of discriminative personality representations.

To alleviate this issue, we propose a Distribution-Calibrated Personality Regression (DCPR) strategy. Given an original personality score $y$, we first apply the Yeo-Johnson transformation
\begin{equation}
T_{\mathrm{YJ}}(y;\lambda)=
\begin{cases}
\frac{(y+1)^\lambda-1}{\lambda}, & \lambda \neq 0,\; \\
\log(y+1), & \lambda = 0,\; \\
\end{cases}
\end{equation}
where $\lambda$ controls the shape of the transformation. For each personality dimension, $\lambda$ is automatically estimated from the training labels by maximum likelihood, rather than manually specified. The transformation reduces distribution skewness and stabilizes variance while preserving the relative ordering of personality scores. After the transformation, Gaussian kernel smoothing is applied to further regularize the target distribution and reduce the influence of annotation noise. The resulting calibrated labels are subsequently used as regression targets during training.

The calibration procedure is performed independently for each personality dimension. During inference, predictions are mapped back to the original score space through the inverse Yeo-Johnson transformation. Figure~\ref{fig:label_dist} illustrates the label distributions before and after calibration, showing that the proposed strategy produces smoother and more balanced targets for personality regression.

\section{Experiments}
\label{sec:experiments}
To answer the central question of this work—whether all modalities contribute equally to different personality traits—we conduct extensive experiments on the AVI Challenge 2026 dataset. We evaluate the effectiveness of different foundation models, modality combinations, fusion mechanisms, and label calibration strategies, with particular emphasis on trait-specific modality preferences. This section introduces the dataset and evaluation protocol, describes the experimental settings, presents ablation studies, and compares our approach with other challenge submissions.

\subsection{Dataset}

We conduct experiments on the official AVI Challenge 2026 dataset, which contains structured video interviews from 644 participants and is split into training, validation, and test sets with a ratio of 7:1:2. Each participant answers six questions, including two general questions and four questions related to HEXACO personality traits. The dataset provides personality scores ranging from 1 to 5, along with demographic metadata that include age, sex, education, and work experience. Model performance is evaluated by the average Mean Squared Error (MSE) across the four personality dimensions. During training, we merge the official training and validation sets and perform five-fold cross-validation. The reported validation performance is computed as the average MSE across the five folds, and the final prediction is obtained by averaging the outputs of the five trained models.

\subsection{Experimental Setup}

During the feature extraction stage, visual, acoustic, and personality\-oriented textual representations are extracted using four NVIDIA RTX 4090 GPUs, each equipped with 24GB memory. All model training and ablation studies are conducted on two NVIDIA RTX 3090 GPUs with 24GB memory per GPU. 

The model is optimized using AdamW with a learning rate of 1e-4 and a batch size of 16. Hyperparameters are selected via grid search, including the dropout rate of the regression head, the complexity of the regression network, and the number of attention heads in attention-based fusion modules. All models are trained for 200 epochs, and for each fold, the checkpoint with the lowest validation MSE on the held-out split is retained for evaluation. Each personality trait is optimized independently through its own regression pathway and model selection process.

\subsection{Ablation Study}

\textbf{Foundation representation models.}
We first compare different foundation representation models under single-modality settings to evaluate their ability to capture personality-relevant information from audio, video, and text inputs. Since the proposed framework ultimately combines multiple modalities, it is important to first assess the quality of modality-specific representations. To isolate the effect of feature extraction quality from downstream fusion mechanisms, each modality is evaluated independently.

For audio-only features, we compare Whisper-base~\cite{pmlr-v202-radford23a}, Emotion2Vec~\cite{ma-etal-2024-emotion2vec}, Gemini Embedding~\cite{lee2025geminiembeddinggeneralizableembeddings}, and Gemini Embedding 2~\cite{shanbhogue2026geminiembedding2native}; for video-only features, ViTMAE~\cite{He2021MaskedAA}, CLIP~\cite{Radford2021LearningTV}, Gemini Embedding, and Gemini Embedding 2 are evaluated; for text-only features, we include BERT-large~\cite{devlin-etal-2019-bert}, RoBERTa-base~\cite{liu2019robertarobustlyoptimizedbert}, Flan-T5-large~\cite{10.5555/3722577.3722647}, text-embedding-3-small~\cite{text_embedding_3_small}, and Qwen3-Embedding-8B~\cite{qwen3embedding}. Within each group, all models follow the same regression and label calibration setting. As shown in Table~\ref{tab:backbone}, the reported MSE is averaged across five folds. Gemini Embedding 2 consistently achieves the best performance across all three modality groups. Compared with other representative foundation models, it obtains the lowest average MSE and exhibits stronger capability in capturing personality-relevant semantic and behavioral signals. This result suggests that a unified multimodal embedding space is more effective for personality representation learning than modality-specific feature extractors. Therefore, Gemini Embedding 2 is adopted as the default feature extractor in the subsequent experiments.

\begin{table}[t]
\centering
\caption{Ablation results of different foundation representation models under single-modality settings. Results are reported as five-fold average MSE. $^\ast$ denotes LLM-based models.}
\label{tab:backbone}
\resizebox{\linewidth}{!}{
\begin{tabular}{lccccc}
\toprule
Foundation Model & H $\downarrow$ & E $\downarrow$ & A $\downarrow$ & C $\downarrow$ & Avg. $\downarrow$ \\
\midrule
\rowcolor{gray!15}
\multicolumn{6}{l}{\textit{Audio-only}} \\
Whisper-base~\cite{pmlr-v202-radford23a} & 0.2536 & 0.4465 & 0.3637 & 0.2522 & 0.3290 \\
Emotion2Vec+ base~\cite{ma-etal-2024-emotion2vec} & 0.2297 & 0.3398 & 0.3527 & 0.2274 & 0.2874 \\
Gemini Embedding$^\ast$~\cite{lee2025geminiembeddinggeneralizableembeddings} & 0.2194 & 0.3528 & 0.3396 & 0.1917 & 0.2759 \\
Gemini Embedding 2$^\ast$~\cite{shanbhogue2026geminiembedding2native} & \textbf{0.2121} & \textbf{0.3337} & \textbf{0.3310} & \textbf{0.1869} & \textbf{0.2659} \\
\midrule
\rowcolor{gray!15}
\multicolumn{6}{l}{\textit{Video-only}} \\
ViTMAE~\cite{He2021MaskedAA} & 0.2616 & 0.4592 & 0.3823 & 0.2646 & 0.3419 \\
CLIP~\cite{Radford2021LearningTV} & 0.3106 & 0.4377 & 0.4155 & 0.3356 & 0.3749 \\
Gemini Embedding$^\ast$~\cite{lee2025geminiembeddinggeneralizableembeddings} & 0.2176 & 0.3319 & 0.3718 & 0.2527 & 0.2935 \\
Gemini Embedding 2$^\ast$~\cite{shanbhogue2026geminiembedding2native} & \textbf{0.2083} & \textbf{0.3182} & \textbf{0.3595} & \textbf{0.2394} & \textbf{0.2814} \\
\midrule
\rowcolor{gray!15}
\multicolumn{6}{l}{\textit{Text-only}} \\
BERT-large$^\ast$~\cite{devlin-etal-2019-bert} & 0.2538 & 0.3657 & 0.3442 & 0.2316 & 0.2988 \\
RoBERTa-base$^\ast$~\cite{liu2019robertarobustlyoptimizedbert} & 0.4019 & 0.6341 & 0.4456 & 0.3531 & 0.4587 \\
Flan-T5-large$^\ast$~\cite{10.5555/3722577.3722647} & 0.2398 & 0.3491 & 0.3290 & 0.2186 & 0.2841 \\
text-embedding-3-small$^\ast$~\cite{text_embedding_3_small} & 0.2319 & 0.3420 & 0.3215 & 0.2062 & 0.2754 \\
Qwen3-Embedding-8B$^\ast$~\cite{qwen3embedding} & 0.2264 & 0.3346 & 0.3120 & \textbf{0.2011} & 0.2685 \\
Gemini Embedding$^\ast$~\cite{lee2025geminiembeddinggeneralizableembeddings} & 0.2275 & 0.3412 & 0.3196 & 0.2071 & 0.2734 \\
Gemini Embedding 2$^\ast$~\cite{shanbhogue2026geminiembedding2native} & \textbf{0.2228} & \textbf{0.3253} & \textbf{0.3078} & 0.2053 & \textbf{0.2653} \\
\bottomrule
\end{tabular}
}
\end{table}

\textbf{Modality combinations.}
We investigate the contribution of different modality combinations using Gemini Embedding 2 features. To isolate modality selection from fusion design, single-modality settings use modality-specific regression, while multimodal settings use feature concatenation followed by the same regression architecture. Hyperparameters are optimized by grid search for each setting. As shown in Table~\ref{tab:modality}, different traits show distinct modality preferences: Honesty-Humility performs best with text and audio, Extraversion with video, Agreeableness with text, and Conscientiousness with all three modalities. No single modality combination achieves the best result across all dimensions. By constructing a trait-level modality configuration, the average MSE is reduced from 0.2631, obtained by the best fixed multimodal setting, to 0.2521. These results indicate that modalities are not equally informative across personality traits.

\begin{table}[t]
\centering
\caption{Ablation results of different modality combinations.}
\label{tab:modality}
\resizebox{\linewidth}{!}{
\begin{tabular}{lccccc}
\toprule
Modality & H $\downarrow$ & E $\downarrow$ & A $\downarrow$ & C $\downarrow$ & Avg. $\downarrow$ \\
\midrule
Text-Only    & 0.2228          & 0.3253          & \textbf{0.3078} & 0.2053          & 0.2653  \\
Audio-Only   & 0.2121          & 0.3337          & 0.3310          & 0.1869          & 0.2659 \\
Video-Only   & 0.2083          & 0.3182 & 0.3595          & 0.2394          & 0.2814 \\
Text + Audio & \textbf{0.1983} & 0.3435          & 0.3295          & 0.1852          & 0.2641 \\
Text + Video & 0.2122          & 0.3406          & 0.3355          & 0.1894          & 0.2694 \\
Text + Audio + Video & 0.2042  & 0.3293          & 0.3349          & \textbf{0.1839} & 0.2631 \\
\midrule
Trait-specific (valid) & \textbf{0.1983} & 0.3182 & \textbf{0.3078} & \textbf{0.1839} & \textbf{0.2521} \\
Trait-specific (test) & 0.3079 & \textbf{0.2886} & 0.3273 & 0.1868 & 0.2777 \\
\bottomrule
\end{tabular}
}
\end{table}

\textbf{Fusion strategies.}

We further compare fusion strategies under the selected modality setting of each personality dimension. Since Extraversion and Agreeableness achieve their best results with a single modality, no cross-modal fusion is required for these two traits, and their best single-modality results are reported as references. For Honesty-Humility and Conscientiousness, which use multimodal inputs, we compare concatenation, cross-modal attention, and attention pooling. As shown in Table~\ref{tab:fusion}, Honesty-Humility performs best with concatenation, while Conscientiousness obtains the lowest MSE with cross-modal attention. This suggests that fusion effectiveness is also trait-dependent, especially when different traits require different forms of cross-modal interaction.

\begin{table}[t]
\centering
\caption{Ablation results of different fusion strategies.}
\label{tab:fusion}
\begin{tabular}{lcccc}
\toprule
Fusion Strategy & H $\downarrow$ & E $\downarrow$ & A $\downarrow$ & C $\downarrow$ \\
\midrule
Concatenation & \textbf{0.1983} & - & - & 0.1839  \\
Cross-Modal Attention & 0.2039 & - & - & \textbf{0.1806} \\
Attention Pooling & 0.2133 & - & - & 0.1955 \\
Best Single Modality & - & \textbf{0.3182} & \textbf{0.3078} & - \\
\bottomrule
\end{tabular}
\end{table}

\textbf{Final trait configuration.}
Based on the above modality and fusion ablations, we construct the final trait-level configuration shown in Table~\ref{tab:trait_config}. Honesty-Humility uses text and audio with attention pooling, Extraversion uses the visual pathway, Agreeableness uses the textual pathway, and Conscientiousness combines all modalities through cross-modal attention. This configuration is used in the final model and the following distribution calibration experiment.

\begin{table}[t]
\centering
\caption{Trait-level modality and fusion configuration used in the final model.}
\label{tab:trait_config}
\resizebox{0.9\linewidth}{!}{
\begin{tabular}{lll}
\toprule
Trait & Modality & Fusion Strategy \\
\midrule
H & Text + Audio & Concatenation \\
E & Video & Modality-Specific \\
A & Text & Modality-Specific \\
C & Text + Audio + Video & Cross-Modal Attention \\
\bottomrule
\end{tabular}
}
\end{table}

\textbf{Distribution calibration and ensemble prediction.}
To evaluate the effectiveness of Distribution-Calibrated Personality Regression, we compare models trained with raw labels and calibrated labels under the same feature and fusion settings. After applying Yeo-Johnson transformation and Gaussian smoothing, the five-fold average MSE decreases from 0.2593 to 0.2521, indicating that target calibration helps alleviate the central tendency bias of personality annotations and provides more stable regression supervision.

\subsection{Comparison with Challenge Submissions}

Our final model is evaluated in the testing phase of the AVI Challenge 2026 Personality Assessment Track. Table~\ref{tab:test_result} reports the official top-10 leaderboard results. Our team (avi\_team) achieves the lowest test MSE of 0.27767 and ranks first among all submissions. This further confirms the effectiveness of trait-specific modality selection and asymmetric fusion in multimodal personality assessment.

\begin{table}[t]
\centering
\caption{Top-10 results on the official test set of the AVI Challenge 2026 Personality Assessment Track.}
\label{tab:test_result}
\resizebox{\linewidth}{!}{
\begin{tabular}{clcc}
\toprule
Rank & Participant & Submission ID & MSE $\downarrow$ \\
\midrule
\textbf{1st} & \textbf{avi\_team} & \textbf{730232} & \textbf{0.27767} \\
2nd & visionxl & 730237 & 0.27768 \\
3rd & hitszly & 729621 & 0.27796 \\
4th & zcst\_avi\_team & 730202 & 0.31035 \\
5th & xiaomaomi & 730144 & 0.32290 \\
6th & CASIA-AVI Team & 729357 & 0.32459 \\
7th & yuyu7879 & 725854 & 0.33648 \\
8th & saunak & 718648 & 0.33703 \\
9th & coderz & 718643 & 0.34231 \\
10th & 	ningsun & 718730 & 0.34455 \\
\bottomrule
\end{tabular}
}
\end{table}

\section{Conclusion}
This paper investigated whether different modalities are equally informative for multimodal personality assessment from a trait-specific perspective. Experimental results show that different HEXACO personality dimensions exhibit distinct modality preferences, supporting our validation-guided trait-specific modeling strategy. These findings challenge the common assumption that a unified multimodal architecture is equally suitable for all personality traits. Based on this observation, we developed a trait-specific asymmetric fusion framework that enables each personality dimension to use its most informative modality combination and fusion pathway. Experiments on the AVI Challenge 2026 dataset demonstrate the effectiveness of the proposed approach, leading to clear improvements over uniform fusion settings and competitive performance in the official evaluation. Future work will explore adaptive instance-level modality selection and more interpretable fusion mechanisms to better understand how behavioral signals contribute to personality assessment.

\begin{acks}
This work was supported by the National Key Research and Development Program of China under Grant No. 2023YFC2506803. This work was also partially supported by the Fundamental Research Funds for the Central Universities under Grant No. PA2025IISL0110. The computations were performed on the HPC Platform of Hefei University of Technology.
\end{acks}

\bibliographystyle{ACM-Reference-Format}
\balance
\bibliography{sample-base}

@String{Computing = "Computing" }

@String{Computer = "{IEEE} Computer" }

@String{Springer = "Springer-Verlag" }

@article{Barrick1991THEBF,
  title={THE BIG FIVE PERSONALITY DIMENSIONS AND JOB PERFORMANCE: A META-ANALYSIS},
  author={Murray R. Barrick and Michael K. Mount},
  journal={Personnel Psychology},
  year={1991},
  volume={44},
  pages={1-26},
  url={https://api.semanticscholar.org/CorpusID:144689146}
}

@article{Paunonen2001BigFF,
  title={Big five factors and facets and the prediction of behavior.},
  author={S. V. Paunonen and Michael C Ashton},
  journal={Journal of personality and social psychology},
  year={2001},
  volume={81 3},
  pages={
          524-39
        },
  url={https://api.semanticscholar.org/CorpusID:30756422}
}

@article{10.1109/TAFFC.2015.2513401,
author = {Celiktutan, Oya and Gunes, Hatice},
title = {Automatic Prediction of Impressions in Time and across Varying Context: Personality, Attractiveness and Likeability},
year = {2017},
issue_date = {January 2017},
publisher = {IEEE Computer Society Press},
address = {Washington, DC, USA},
volume = {8},
number = {1},
issn = {1949-3045},
url = {https://doi.org/10.1109/TAFFC.2015.2513401},
doi = {10.1109/TAFFC.2015.2513401},
journal = {IEEE Trans. Affect. Comput.},
month = jan,
pages = {29–42},
numpages = {14}
}

@article{Vinciarelli2014ASO,
  title={A Survey of Personality Computing},
  author={Alessandro Vinciarelli and Gelareh Mohammadi},
  journal={IEEE Transactions on Affective Computing},
  year={2014},
  volume={5},
  pages={273-291},
  url={https://api.semanticscholar.org/CorpusID:14688441}
}

@inproceedings{PonceLpez2016ChaLearnL2,
  title={ChaLearn LAP 2016: First Round Challenge on First Impressions - Dataset and Results},
  author={V{\'i}ctor Ponce-L{\'o}pez and Baiyu Chen and Marc Oliu and Ciprian Adrian Corneanu and Albert Clap{\'e}s and Isabelle M Guyon and Xavier Bar{\'o} and Hugo Jair Escalante and Sergio Escalera},
  booktitle={ECCV Workshops},
  year={2016},
  url={https://api.semanticscholar.org/CorpusID:39164174}
}

@inproceedings{Zhang2016DeepBR,
  title={Deep Bimodal Regression for Apparent Personality Analysis},
  author={Chen-Lin Zhang and Hao Zhang and Xiu-Shen Wei and Jianxin Wu},
  booktitle={ECCV Workshops},
  year={2016},
  url={https://api.semanticscholar.org/CorpusID:26959563}
}

@article{Escalante2017DesignOA,
  title={Design of an explainable machine learning challenge for video interviews},
  author={Hugo Jair Escalante and Isabelle M Guyon and Sergio Escalera and Julio C. S. Jacques Junior and Meysam Madadi and Xavier Bar{\'o} and S. Ayache and Evelyne Viegas and Yağmur G{\"u}çl{\"u}t{\"u}rk and Umut G{\"u}çl{\"u} and Marcel van Gerven and Robert van Lier},
  journal={2017 International Joint Conference on Neural Networks (IJCNN)},
  year={2017},
  pages={3688-3695},
  url={https://api.semanticscholar.org/CorpusID:22530140}
}

@article{Aslan2021MultimodalAO,
  title={Multimodal assessment of apparent personality using feature attention and error consistency constraint},
  author={S{\"u}leyman Aslan and Uğur G{\"u}d{\"u}kbay and Hamdi Dibeklioğlu},
  journal={Image Vis. Comput.},
  year={2021},
  volume={110},
  pages={104163},
  url={https://api.semanticscholar.org/CorpusID:233657208}
}

@article{10.1109/TAFFC.2024.3363710,
author = {Liao, Rongfan and Song, Siyang and Gunes, Hatice},
title = {An Open-Source Benchmark of Deep Learning Models for Audio-Visual Apparent and Self-Reported Personality Recognition},
year = {2024},
issue_date = {July-Sept. 2024},
publisher = {IEEE Computer Society Press},
address = {Washington, DC, USA},
volume = {15},
number = {3},
issn = {1949-3045},
url = {https://doi.org/10.1109/TAFFC.2024.3363710},
doi = {10.1109/TAFFC.2024.3363710},
journal = {IEEE Trans. Affect. Comput.},
month = jul,
pages = {1590–1607},
numpages = {18}
}

@article{Lee2024GeckoVT,
  title={Gecko: Versatile Text Embeddings Distilled from Large Language Models},
  author={Jinhyuk Lee and Zhuyun Dai and Xiaoqi Ren and Blair Chen and Daniel Cer and Jeremy R. Cole and Kai Hui and Michael Boratko and Rajvi Kapadia and Wen Ding and Yi Luan and Sai Meher Karthik Duddu and Gustavo Hern{\'a}ndez Abrego and Wei Shi and Nithi Gupta and Aditya Kusupati and Prateek Jain and Siddhartha R. Jonnalagadda and Ming-Wei Chang and Iftekhar Naim},
  journal={ArXiv},
  year={2024},
  volume={abs/2403.20327},
  url={https://api.semanticscholar.org/CorpusID:268793455}
}

@article{john1999big,
  title={The Big-Five trait taxonomy: History, measurement, and theoretical perspectives},
  author={John, Oliver},
  journal={Published as},
  year={1999}
}

@article{ashton2009hexaco,
  title={The HEXACO--60: A short measure of the major dimensions of personality},
  author={Ashton, Michael C and Lee, Kibeom},
  journal={Journal of personality assessment},
  volume={91},
  number={4},
  pages={340--345},
  year={2009},
  publisher={Taylor \& Francis}
}

@inproceedings{ponce2016chalearn,
  title={Chalearn lap 2016: First round challenge on first impressions-dataset and results},
  author={Ponce-L{\'o}pez, V{\'\i}ctor and Chen, Baiyu and Oliu, Marc and Corneanu, Ciprian and Clap{\'e}s, Albert and Guyon, Isabelle and Bar{\'o}, Xavier and Escalante, Hugo Jair and Escalera, Sergio},
  booktitle={European conference on computer vision},
  pages={400--418},
  year={2016},
  organization={Springer}
}

@inproceedings{li2025traits,
  title={Traits Run Deep: Enhancing Personality Assessment via Psychology-Guided LLM Representations and Multimodal Apparent Behaviors},
  author={Li, Jia and He, Yichao and Xu, Jiacheng and Luo, Tianhao and Hu, Zhenzhen and Hong, Richang and Wang, Meng},
  booktitle={Proceedings of the 33rd ACM International Conference on Multimedia},
  pages={13901--13908},
  year={2025}
}

@inproceedings{cui2025less,
  title={Less is More? Textual-Only Language Model for AVI challenge 2025},
  author={Cui, Jizhou and Xu, Hanzhe and Liu, Xuefei and Lian, Zheng and Wen, Zhengqi and Xie, Heng and Fu, Ruibo and Liu, Yukun and Tao, Jianhua},
  booktitle={Proceedings of the 3rd International Workshop on Multimodal and Responsible Affective Computing},
  pages={69--74},
  year={2025}
}

@inproceedings{yang2025enhancing,
  title={Enhancing Multimodal Personality Assessment with LLM-Augmented Hierarchical Fusion},
  author={Yang, Longjiang and Yu, Cong and Huang, Chenxi and Zhang, Fengyu and Liu, Ran and Wen, Zhuofan and Chen, Shun and Yao, Hailiang and Liu, Bin and Lian, Zheng and others},
  booktitle={Proceedings of the 33rd ACM International Conference on Multimedia},
  pages={13917--13923},
  year={2025}
}

@article{zhangavi,
  title={AVI Challenge 2026: Assessing True Personality Traits and Cognitive Ability from Asynchronous Video Interviews (AVIs)},
  author={Zhang, Tianyi and Qi, Tianhua and de Vries, Reinout E and ZHENG, Wenming and Oostrom, Janneke and Holtrop, Djurre and Zong, Yuan and Koutsoumpis, Antonios}
}

@article{ryumina2024gated,
  title={Gated Siamese Fusion Network based on multimodal deep and hand-crafted features for personality traits assessment},
  author={Ryumina, Elena and Markitantov, Maxim and Ryumin, Dmitry and Karpov, Alexey},
  journal={Pattern Recognition Letters},
  volume={185},
  pages={45--51},
  year={2024},
  publisher={Elsevier}
}

@inproceedings{masumura2025multimodal,
  title={Multimodal fine-grained apparent personality trait recognition: Joint modeling of big five and questionnaire item-level scores},
  author={Masumura, Ryo and Orihashi, Shota and Ihori, Mana and Tanaka, Tomohiro and Makishima, Naoki and Suzuki, Satoshi and Mizuno, Saki and Hojo, Nobukatsu},
  booktitle={Proceedings of the AAAI Conference on Artificial Intelligence},
  volume={39},
  number={2},
  pages={1456--1464},
  year={2025}
}

@article{zhao2023integrating,
  title={Integrating audio and visual modalities for multimodal personality trait recognition via hybrid deep learning},
  author={Zhao, Xiaoming and Liao, Yuehui and Tang, Zhiwei and Xu, Yicheng and Tao, Xin and Wang, Dandan and Wang, Guoyu and Lu, Hongsheng},
  journal={Frontiers in Neuroscience},
  volume={16},
  pages={1107284},
  year={2023},
  publisher={Frontiers Media SA}
}

@inproceedings{devlin-etal-2019-bert,
    title = "{BERT}: Pre-training of Deep Bidirectional Transformers for Language Understanding",
    author = "Devlin, Jacob  and
      Chang, Ming-Wei  and
      Lee, Kenton  and
      Toutanova, Kristina",
    editor = "Burstein, Jill  and
      Doran, Christy  and
      Solorio, Thamar",
    booktitle = "Proceedings of the 2019 Conference of the North {A}merican Chapter of the Association for Computational Linguistics: Human Language Technologies, Volume 1 (Long and Short Papers)",
    month = jun,
    year = "2019",
    address = "Minneapolis, Minnesota",
    publisher = "Association for Computational Linguistics",
    url = "https://aclanthology.org/N19-1423/",
    doi = "10.18653/v1/N19-1423",
    pages = "4171--4186",
}

@misc{liu2019robertarobustlyoptimizedbert,
      title={RoBERTa: A Robustly Optimized BERT Pretraining Approach}, 
      author={Yinhan Liu and Myle Ott and Naman Goyal and Jingfei Du and Mandar Joshi and Danqi Chen and Omer Levy and Mike Lewis and Luke Zettlemoyer and Veselin Stoyanov},
      year={2019},
      eprint={1907.11692},
      archivePrefix={arXiv},
      primaryClass={cs.CL},
      url={https://arxiv.org/abs/1907.11692}, 
}

@article{10.5555/3722577.3722647,
author = {Chung, Hyung Won and Hou, Le and Longpre, Shayne and Zoph, Barret and Tai, Yi and Fedus, William and Li, Yunxuan and Wang, Xuezhi and Dehghani, Mostafa and Brahma, Siddhartha and Webson, Albert and Gu, Shixiang Shane and Dai, Zhuyun and Suzgun, Mirac and Chen, Xinyun and Chowdhery, Aakanksha and Castro-Ros, Alex and Pellat, Marie and Robinson, Kevin and Valter, Dasha and Narang, Sharan and Mishra, Gaurav and Yu, Adams and Zhao, Vincent and Huang, Yanping and Dai, Andrew and Yu, Hongkun and Petrov, Slav and Chi, Ed H. and Dean, Jeff and Devlin, Jacob and Roberts, Adam and Zhou, Denny and Le, Quoc V. and Wei, Jason},
title = {Scaling instruction-finetuned language models},
year = {2024},
issue_date = {January 2024},
publisher = {JMLR.org},
volume = {25},
number = {1},
issn = {1532-4435},
journal = {J. Mach. Learn. Res.},
month = jan,
articleno = {70},
numpages = {53}
}

@article{qwen3embedding,
  title={Qwen3 Embedding: Advancing Text Embedding and Reranking Through Foundation Models},
  author={Zhang, Yanzhao and Li, Mingxin and Long, Dingkun and Zhang, Xin and Lin, Huan and Yang, Baosong and Xie, Pengjun and Yang, An and Liu, Dayiheng and Lin, Junyang and Huang, Fei and Zhou, Jingren},
  journal={arXiv preprint arXiv:2506.05176},
  year={2025}
}

@misc{lee2025geminiembeddinggeneralizableembeddings,
      title={Gemini Embedding: Generalizable Embeddings from Gemini}, 
      author={Jinhyuk Lee and Feiyang Chen and Sahil Dua and Daniel Cer and Madhuri Shanbhogue and Iftekhar Naim and Gustavo Hernández Ábrego and Zhe Li and Kaifeng Chen and Henrique Schechter Vera and Xiaoqi Ren and Shanfeng Zhang and Daniel Salz and Michael Boratko and Jay Han and Blair Chen and Shuo Huang and Vikram Rao and Paul Suganthan and Feng Han and Andreas Doumanoglou and Nithi Gupta and Fedor Moiseev and Cathy Yip and Aashi Jain and Simon Baumgartner and Shahrokh Shahi and Frank Palma Gomez and Sandeep Mariserla and Min Choi and Parashar Shah and Sonam Goenka and Ke Chen and Ye Xia and Koert Chen and Sai Meher Karthik Duddu and Yichang Chen and Trevor Walker and Wenlei Zhou and Rakesh Ghiya and Zach Gleicher and Karan Gill and Zhe Dong and Mojtaba Seyedhosseini and Yunhsuan Sung and Raphael Hoffmann and Tom Duerig},
      year={2025},
      eprint={2503.07891},
      archivePrefix={arXiv},
      primaryClass={cs.CL},
      url={https://arxiv.org/abs/2503.07891}, 
}

@misc{shanbhogue2026geminiembedding2native,
      title={Gemini Embedding 2: A Native Multimodal Embedding Model from Gemini}, 
      author={Madhuri Shanbhogue and Zhe Li and Shanfeng Zhang and Gustavo Hernández Ábrego and Shih-Cheng Huang and Aashi Jain and Daniel Salz and Sonam Goenka and Chaitra Hegde and Ji Ma and Feiyang Chen and Jiaxing Wu and Tanmaya Dabral and Babak Samari and Kevin Poulet and Daniel Cer and Kaifeng Chen and Paul Suganathan and Hui Hui and Jovan Andonov and Philippe Schlattner and Jay Han and Iftekhar Naim and Wing Lowe and Vladimir Pchelin and Albert Yang and Yi-Ting Chen and Zhongli Ding and Grace Zhang and Georg Heigold and Yichang Chen and Antoine Reveillon and Brendan Mccloskey and Wenlei Zhou and Dahun Kim and Rui Meng and Emma Wang and Jack Zheng and Halley Fede and Zhen Yang and Keegan Mosley and Brian Potetz and Sahil Dua and Henrique Schechter Vera and Shen Gao and Hesen Zhang and Andreas Hess and Hengxuan Ying and Alberto Montes and Karan Gill and Min Choi and Sebastian Russo and Anja Hauth and Jinhyuk Lee and Michael Boratko and Megan Barnes and Vikram Rao and Claudiu Musat and Cyril Allauzen and Ehsan Variani and Shankar Kumar and Tom Bagby and Junyi Jiao and Yang Gu and Tengxin Li and Ayush Agrawal and Roberto Santana and Dev Nath and Stephen Karukas and Shuoxuan Han and Lucia Loher and Alice Twu and Nidhi Vyas and Siddharth Bhai and Frank Palma Gomez and Wangyuan Zhang and Chaoren Liu and Jizheng Yang and Steve Qiu and Shijie Zhang and Sujay Kulkarni and Sascha Rothe and Sean Nakamoto and Raphael Hoffmann and Zach Gleicher and Yunhsuan Sung and Qin Yin and Tom Duerig and Mojtaba Seyedhosseini},
      year={2026},
      eprint={2605.27295},
      archivePrefix={arXiv},
      primaryClass={cs.CV},
      url={https://arxiv.org/abs/2605.27295}, 
}

@inproceedings{ma-etal-2024-emotion2vec,
    title = "emotion2vec: Self-Supervised Pre-Training for Speech Emotion Representation",
    author = "Ma, Ziyang  and
      Zheng, Zhisheng  and
      Ye, Jiaxin  and
      Li, Jinchao  and
      Gao, Zhifu  and
      Zhang, ShiLiang  and
      Chen, Xie",
    editor = "Ku, Lun-Wei  and
      Martins, Andre  and
      Srikumar, Vivek",
    booktitle = "Findings of the Association for Computational Linguistics: ACL 2024",
    month = aug,
    year = "2024",
    address = "Bangkok, Thailand",
    publisher = "Association for Computational Linguistics",
    url = "https://aclanthology.org/2024.findings-acl.931/",
    doi = "10.18653/v1/2024.findings-acl.931",
    pages = "15747--15760",
}

@InProceedings{pmlr-v202-radford23a,
  title = 	 {Robust Speech Recognition via Large-Scale Weak Supervision},
  author =       {Radford, Alec and Kim, Jong Wook and Xu, Tao and Brockman, Greg and Mcleavey, Christine and Sutskever, Ilya},
  booktitle = 	 {Proceedings of the 40th International Conference on Machine Learning},
  pages = 	 {28492--28518},
  year = 	 {2023},
  editor = 	 {Krause, Andreas and Brunskill, Emma and Cho, Kyunghyun and Engelhardt, Barbara and Sabato, Sivan and Scarlett, Jonathan},
  volume = 	 {202},
  series = 	 {Proceedings of Machine Learning Research},
  month = 	 {23--29 Jul},
  publisher =    {PMLR},
  url = 	 {https://proceedings.mlr.press/v202/radford23a.html}
}

@article{He2021MaskedAA,
  title={Masked Autoencoders Are Scalable Vision Learners},
  author={Kaiming He and Xinlei Chen and Saining Xie and Yanghao Li and Piotr Doll'ar and Ross B. Girshick},
  journal={2022 IEEE/CVF Conference on Computer Vision and Pattern Recognition (CVPR)},
  year={2021},
  pages={15979-15988},
  url={https://api.semanticscholar.org/CorpusID:243985980}
}

@article{Radford2021LearningTV,
  title={Learning Transferable Visual Models From Natural Language Supervision},
  author={Alec Radford and Jong Wook Kim and Chris Hallacy and Aditya Ramesh and Gabriel Goh and Sandhini Agarwal and Girish Sastry and Amanda Askell and Pamela Mishkin and Jack Clark and Gretchen Krueger and Ilya Sutskever},
  journal={ArXiv},
  year={2021},
  volume={abs/2103.00020},
  url={https://api.semanticscholar.org/CorpusID:231591445}
}

@article{10.1093/biomet/87.4.954,
    author = {Yeo, In‐Kwon and Johnson, Richard A.},
    title = {A new family of power transformations to improve normality or symmetry},
    journal = {Biometrika},
    volume = {87},
    number = {4},
    pages = {954-959},
    year = {2000},
    month = {12},
    issn = {0006-3444},
    doi = {10.1093/biomet/87.4.954},
    url = {https://doi.org/10.1093/biomet/87.4.954},
    eprint = {https://academic.oup.com/biomet/article-pdf/87/4/954/633221/870954.pdf},
}

@article{Pham2018FoundIT,
  title={Found in Translation: Learning Robust Joint Representations by Cyclic Translations Between Modalities},
  author={Hai Pham and Paul Pu Liang and Thomas Manzini and Louis-philippe Morency and Barnab{\'a}s P{\'o}czos},
  journal={ArXiv},
  year={2018},
  volume={abs/1812.07809},
  url={https://api.semanticscholar.org/CorpusID:53500027}
}

@article{Zhang2022TailorVM,
  title={Tailor Versatile Multi-modal Learning for Multi-label Emotion Recognition},
  author={Yi Zhang and Mingyuan Chen and Jundong Shen and Chongjun Wang},
  journal={ArXiv},
  year={2022},
  volume={abs/2201.05834},
  url={https://api.semanticscholar.org/CorpusID:246015645}
}

@article{Zhao2022DeepPT,
  title={Deep Personality Trait Recognition: A Survey},
  author={Xiaoming Zhao and Zhiwei Tang and Shiqing Zhang},
  journal={Frontiers in Psychology},
  year={2022},
  volume={13},
  url={https://api.semanticscholar.org/CorpusID:248530069}
}

@article{Peters2023LargeLM,
  title={Large language models can infer psychological dispositions of social media users},
  author={Heinrich Peters and Sandra C. Matz},
  journal={PNAS Nexus},
  year={2023},
  volume={3},
  url={https://api.semanticscholar.org/CorpusID:262043671}
}

@Article{electronics14142837,
AUTHOR = {Bhin, Hyeonuk and Choi, Jongsuk},
TITLE = {Multimodal Personality Recognition Using Self-Attention-Based Fusion of Audio, Visual, and Text Features},
JOURNAL = {Electronics},
VOLUME = {14},
YEAR = {2025},
NUMBER = {14},
ARTICLE-NUMBER = {2837},
URL = {https://www.mdpi.com/2079-9292/14/14/2837},
ISSN = {2079-9292},
DOI = {10.3390/electronics14142837}
}

@article{Wang2024ASM,
  title={A single modality apparent first impression personality recognition model with temporal emotion based LSTM},
  author={Jialou Wang and Honglei Li and Wai Lok Woo and Shan Shan},
  journal={Expert Syst. Appl.},
  year={2024},
  volume={259},
  pages={125114},
  url={https://api.semanticscholar.org/CorpusID:272129536}
}

@article{Ning2026ADM,
  title={A DLF multi-scale quantitative research method for the big five personality traits},
  author={Tao Ning and Zhenghua Guo and Qidong Hou},
  journal={Scientific Reports},
  year={2026},
  volume={16},
  url={https://api.semanticscholar.org/CorpusID:284474095}
}

@article{Ren2022BalancedMF,
  title={Balanced MSE for Imbalanced Visual Regression},
  author={Jiawei Ren and Mingyuan Zhang and Cunjun Yu and Ziwei Liu},
  journal={2022 IEEE/CVF Conference on Computer Vision and Pattern Recognition (CVPR)},
  year={2022},
  pages={7916-7925},
  url={https://api.semanticscholar.org/CorpusID:247793481}
}

@article{Zhu2024IRDAID,
  title={IRDA: Implicit data augmentation for deep imbalanced regression},
  author={Weiyao Zhu and Ou Wu and Nan Yang},
  journal={Inf. Sci.},
  year={2024},
  volume={677},
  pages={120873},
  url={https://api.semanticscholar.org/CorpusID:270319009}
}

@article{Puetz2026DeconstructingDI,
  title={Deconstructing deep imbalance regression: a comprehensive review and experimental evaluation},
  author={Noah C. Puetz and Jens U. Brandt and Marc Hilbert and Elena Raponi and Thomas B{\"a}ck and Thomas Bartz-Beielstein},
  journal={Artificial Intelligence Review},
  year={2026},
  url={https://api.semanticscholar.org/CorpusID:287705337}
}

@article{10.5555/1622637.1622649,
author = {Mairesse, Fran\c{c}ois and Walker, Marilyn A. and Mehl, Matthias R. and Moore, Roger K.},
title = {Using linguistic cues for the automatic recognition of personality in conversation and text},
year = {2007},
issue_date = {September 2007},
publisher = {AI Access Foundation},
address = {El Segundo, CA, USA},
volume = {30},
number = {1},
issn = {1076-9757},
journal = {J. Artif. Int. Res.},
month = nov,
pages = {457–500},
numpages = {44}
}

@article{Pennebaker1999LinguisticSL,
  title={Linguistic styles: language use as an individual difference.},
  author={James W. Pennebaker and Laura A. King},
  journal={Journal of personality and social psychology},
  year={1999},
  volume={77 6},
  pages={1296-312},
  url={https://api.semanticscholar.org/CorpusID:29567532}
}

@inproceedings{tan-bansal-2019-lxmert,
    title = "{LXMERT}: Learning Cross-Modality Encoder Representations from Transformers",
    author = "Tan, Hao  and
      Bansal, Mohit",
    editor = "Inui, Kentaro  and
      Jiang, Jing  and
      Ng, Vincent  and
      Wan, Xiaojun",
    booktitle = "Proceedings of the 2019 Conference on Empirical Methods in Natural Language Processing and the 9th International Joint Conference on Natural Language Processing (EMNLP-IJCNLP)",
    month = nov,
    year = "2019",
    address = "Hong Kong, China",
    publisher = "Association for Computational Linguistics",
    url = "https://aclanthology.org/D19-1514/",
    doi = "10.18653/v1/D19-1514",
    pages = "5100--5111",
}

@ARTICLE{8913501,
author={Principi, Ricardo Dario Perez and Palmero, Cristina and Junior, Julio C. S. Jacques and Escalera, Sergio},
journal={ IEEE Transactions on Affective Computing },
title={{ On the Effect of Observed Subject Biases in Apparent Personality Analysis From Audio-Visual Signals }},
year={2021},
volume={12},
number={03},
ISSN={1949-3045},
pages={607-621},
doi={10.1109/TAFFC.2019.2956030},
url = {https://doi.ieeecomputersociety.org/10.1109/TAFFC.2019.2956030},
publisher={IEEE Computer Society},
address={Los Alamitos, CA, USA},
month=jul}

@article{Park2020KEmoConAM,
  title={K-EmoCon, a multimodal sensor dataset for continuous emotion recognition in naturalistic conversations},
  author={Cheul Young Park and Narae Cha and Soowon Kang and Auk Kim and Ahsan Habib Khandoker and Leontios J. Hadjileontiadis and Alice H. Oh and Yong Jeong and Uichin Lee},
  journal={Scientific Data},
  year={2020},
  volume={7},
  url={https://api.semanticscholar.org/CorpusID:218571336}
}

@inproceedings{cheng2025vaemo,
  title={Vaemo: Efficient representation learning for visual-audio emotion with knowledge injection},
  author={Cheng, Hao and Zhao, Zhiwei and He, Yichao and Hu, Zhenzhen and Li, Jia and Wang, Meng and Hong, Richang},
  booktitle={Proceedings of the 33rd ACM International Conference on Multimedia},
  pages={5547--5556},
  year={2025}
}

@ARTICLE{11207542,
  author={Chen, Yin and Li, Jia and Zhang, Yu and Hu, Zhenzhen and Shan, Shiguang and Wang, Meng and Hong, Richang},
  journal={IEEE Transactions on Affective Computing}, 
  title={Static for Dynamic: Towards a Deeper Understanding of Dynamic Facial Expressions Using Static Expression Data}, 
  year={2025},
  volume={},
  number={},
  pages={1-15},
  doi={10.1109/TAFFC.2025.3623135}}

@misc{text_embedding_3_small,
  author = {{OpenAI}},
  year =   {2024},
  title =  {text-embedding-3-small},
  lastaccessed = {June 6, 2026},
  url =    {https://developers.openai.com/api/docs/models/text-embedding-3-small},
}

@inproceedings{10.1145/3746027.3762016,
author = {Zhang, Tianyi and Qi, Tianhua and Koutsoumpis, Antonis and Zong, Yuan and Zheng, Wenming and Oostrom, Janneke K. and Holtrop, Djurre and Luo, Zhaojie and de Vries, Reinout E.},
title = {Assessing Personality Traits and Interview Performance from Asynchronous Video Interviews},
year = {2025},
isbn = {9798400720352},
publisher = {Association for Computing Machinery},
address = {New York, NY, USA},
url = {https://doi.org/10.1145/3746027.3762016},
doi = {10.1145/3746027.3762016},
booktitle = {Proceedings of the 33rd ACM International Conference on Multimedia},
pages = {13895–13900},
numpages = {6},
location = {Dublin, Ireland},
series = {MM '25}
}

@inproceedings{10.1145/3746027.3762064,
author = {Liu, Yuyun and Zhang, Kaifei and Ma, Yinghao and Xu, Xiaolin and Qi, Tianhua and Zheng, Wenming and Lu, Cheng and Zong, Yuan},
title = {Multi-Level Segment Fusion Based on Adaptive Time-Window Selection for Multimodal Personality-Aware Elderly Depression Detection},
year = {2025},
isbn = {9798400720352},
publisher = {Association for Computing Machinery},
address = {New York, NY, USA},
url = {https://doi.org/10.1145/3746027.3762064},
doi = {10.1145/3746027.3762064},
booktitle = {Proceedings of the 33rd ACM International Conference on Multimedia},
pages = {13944–13950},
numpages = {7},
location = {Dublin, Ireland},
series = {MM '25}
}

@article{LI2025113260,
title = {Low-rank joint distribution adaptation for cross-corpus speech emotion recognition},
journal = {Knowledge-Based Systems},
volume = {315},
pages = {113260},
year = {2025},
issn = {0950-7051},
doi = {https://doi.org/10.1016/j.knosys.2025.113260},
url = {https://www.sciencedirect.com/science/article/pii/S0950705125003077},
author = {Sunan Li and Cheng Lu and Yan Zhao and Hailun Lian and Tianhua Qi and Yuan Zong},
}

@article{qi2026affectspeech,
  title        = {AffectSpeech: A Large-Scale Emotional Speech Dataset with Fine-Grained Textual Descriptions for Speech Emotion Captioning and Synthesis},
  author       = {Qi, Tianhua and Zheng, Wenming and Schuller, Bj{\"o}rn W. and Luo, Zhaojie and Li, Haizhou},
  journal={arXiv preprint arXiv:2604.04160},
  year         = {2026}
}

\end{document}